\documentclass[runningheads]{llncs}
\usepackage{graphicx}
\usepackage{amsmath,amssymb} 
\usepackage{color}
\usepackage[width=122mm,left=12mm,paperwidth=146mm,height=193mm,top=12mm,paperheight=217mm]{geometry}
\usepackage{times}
\usepackage{epsfig}
\usepackage{amsmath}
\usepackage{amssymb}
\usepackage{comment}
\usepackage{multirow}
\usepackage[nameinlink, capitalise]{cleveref}	

\newcommand{\IGNORE}[1]{{}}

\newcommand{\VIDEOS}{1004}

\newcommand{\FRAMESSELF}{521,406}
\newcommand{\FRAMESTRAIN}{364,256}
\newcommand{\FRAMESTRAINSELF}{368,135}
\newcommand{\FRAMESVAL}{76,309}
\newcommand{\FRAMESVALSELF}{75,526}
\newcommand{\FRAMESTEST}{78,562}
\newcommand{\FRAMESTESTSELF}{77,745}
\newcommand{\NUMTRAIN}{704}
\newcommand{\NUMVAL}{150}
\newcommand{\NUMTEST}{150}

\begin{document}
\pagestyle{headings}
\mainmatter
\def\ECCV18SubNumber{2151}  

\title{FaceForensics: A Large-scale Video Dataset for Forgery Detection in Human Faces}

\titlerunning{FaceForensics: A Large-scale Video Dataset for Forgery Detection in Human Faces}

\authorrunning{A. R\"ossler, D. Cozzolino, L. Verdoliva, C. Riess \\ J. Thies, M. Nie{\ss}ner}

\author{
	Andreas R\"ossler\textsuperscript{1} \ \ \ 
	Davide Cozzolino\textsuperscript{2} \ \ \ 
	Luisa Verdoliva\textsuperscript{2} \ \ \ 
	Christian Riess\textsuperscript{3}\\
	Justus Thies\textsuperscript{1} \ \ \ \ \ \ 
	Matthias Nie{\ss}ner\textsuperscript{1}
}
\institute{
	\textsuperscript{1}Technical University of Munich \ \ \ \ \ \textsuperscript{2}University Federico II of Naples \\
	\textsuperscript{3}University of Erlangen-Nuremberg
}

\maketitle
	
\begin{center}
	
	\includegraphics[width=1.0\linewidth]{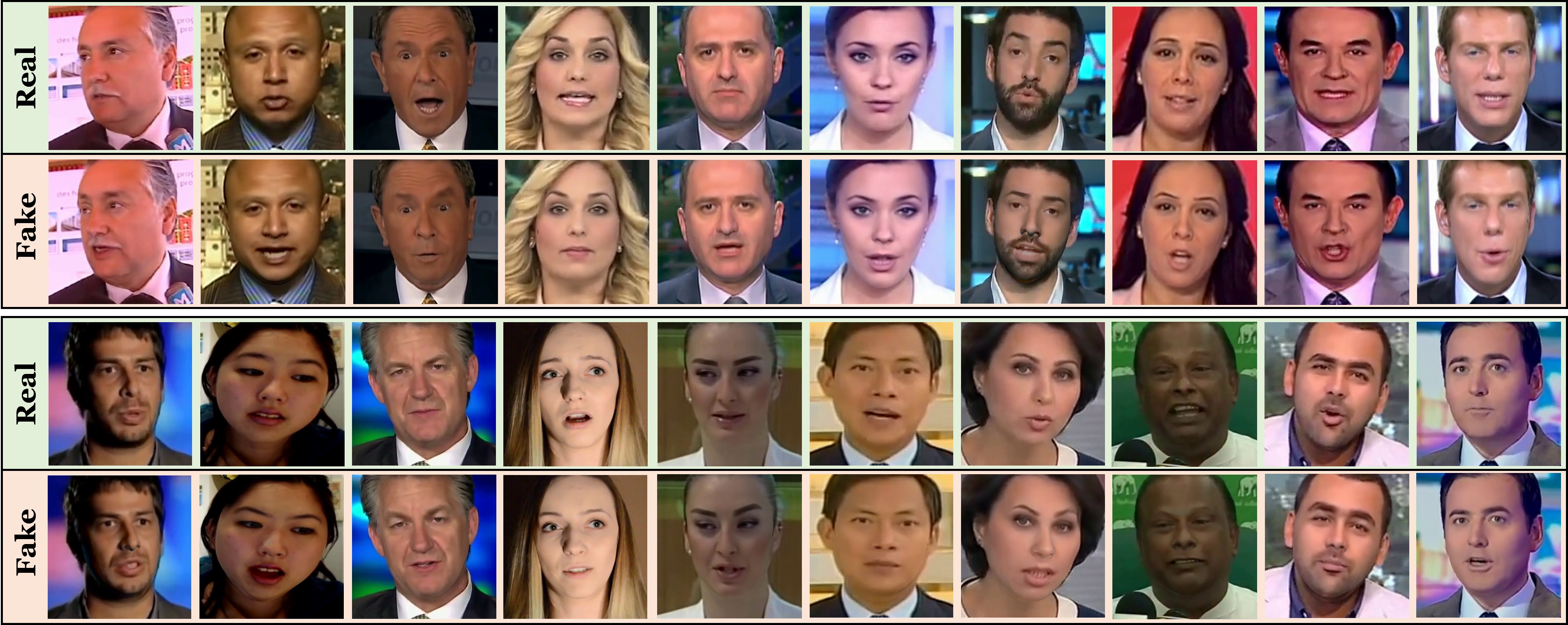}
\end{center}

\begin{abstract}
	With recent advances in computer vision and graphics, it is now possible to generate videos with extremely realistic synthetic faces, even in real time.
Countless applications are possible, some of which raise a legitimate alarm, calling for reliable detectors of fake videos.
In fact, distinguishing between original and manipulated video can be a challenge for humans and computers alike,
especially when the videos are compressed or have low resolution, as it often happens on social networks.
Research on the detection of face manipulations has been seriously hampered by the lack of adequate datasets.
To this end, we introduce a novel face manipulation dataset of about half a million edited images (from over 1000 videos).
The manipulations have been generated with a state-of-the-art face editing approach. It exceeds all existing video manipulation datasets by at least an order of magnitude. 
Using our new dataset, we introduce benchmarks for classical image forensic tasks, including classification and segmentation,
considering videos compressed at various quality levels.
In addition, we introduce a benchmark evaluation for creating indistinguishable forgeries with known ground truth; for instance with generative refinement models.
	
	\begin{keywords}
		Image Forensics, Video Manipulation,  Facial Reenactment
	\end{keywords}
\end{abstract}

\section{Introduction}
Faces play a central role in human interaction, as the face of a person can emphasize a message or it can even convey a message in its own right~\cite{Frith09:RFS}. 
In particular, for faces, we have seen stunning progress in image and video manipulation methods in recent years.
State-of-the-art methods can now generate manipulated videos in real time \cite{Thies16}, can synthesize videos based on audio input \cite{suwajanakorn2017synthesizing}, or can artificially animate static images \cite{elor2017bringingPortraits}.
At the same time, the ability to edit facial expressions has also gained tremendous attention in the context of {fake-news} discussions and in the current political climate in many countries. The ability to effortlessly create visually plausible editing of faces in videos has the potential to severely undermine trust in any form of digital communication. 
For instance, in social networks, filtering out or tagging manipulated images is currently one of the most problematic issues.
Furthermore, the authenticity of face pictures also plays a role in completely different applications, such as biometric access control ~\cite{Jain04:IBR,Ferrara14:TMP}. 

In this context, image forensics research has recently gained momentum in examining the authenticity of images.
Here, we believe the recent advances in deep learning offer a unique opportunity due to the ability to learn extremely powerful image features with convolutional neural networks (CNNs). %
In particular, supervised training has shown to produce extremely impressive results, and we speculate that they could be well-suited to robustly identify manipulations.
Unfortunately, these methods rely on large amounts of training data, and most forensic datasets to date are manually created, thus limited in size.
This lack of available training data is a severe bottleneck for training deep networks for manipulation detection and makes it hard to evaluate different methods.

In order to alleviate this shortage of training samples, we introduce a comprehensive dataset of facial manipulations composed of  
over 500,000 frames from \VIDEOS{} videos using the state-of-the-art Face2Face approach \cite{Thies16}. 
We consider two types of manipulation: \textit{source-to-target}, where we transfer facial expressions from a source video to a target video using Face2Face, and \textit{self-reenactment}, where we use Face2Face to reenact the facial expressions of a source video. 
In addition, we provide the reconstructed face masks generated by Face2Face for all videos in the source-to-target dataset.

Thanks to the \textit{source-to-target} dataset, we can carry out a forensic analysis
and train data reliant algorithms in a realistic scenario, given that the source and target videos were retrieved from YouTube.
In particular, we evaluate the performance of a variety of methods on two main tasks:
forgery \emph{classification} (is anything in an image forged?) and \emph{segmentation} (is the current pixel forged?).
Performance is analyzed on manipulated videos compressed at various quality levels
to account for the typical processing encountered when the video is uploaded on the internet.
This is a very challenging situation since low-level manipulation traces can get lost after compression.

In addition to classification and evaluation, the \textit{self-reenactment} dataset allows us to evaluate generative methods.
In particular, the generation process can start from an already well-structured fake,
which helps us focus on refinement in a possibly supervised environment,
a problem resembling synthetic-to-real translations \cite{shrivastava2017learning}.
Furthermore, the performance of our refinement models can be evaluated using forgery detection approaches, without resorting to subjective metrics, such as visual user studies.
Here, we introduce an evaluation scheme based on creating indistinguishable images based on generative models with known ground truth.
 	
In summary, we introduce two versions of a novel dataset of manipulated facial expressions composed of more than 500,000 images from 1004 videos with pristine sources and target ground truth in \cref{sec:dataset}. 
In particular, our new dataset focuses on the following problem statements:
\begin{itemize}
	\item How well do current state-of-the-art approaches perform in a realistic setting both for forgery detection (\cref{sec:detection}) and segmentation (\cref{sec:segmentation})?
	\item Can we use generative networks to improve the quality of forgeries (\cref{sec:generative})?
\end{itemize}

\section{Related Work}
	\label{sec:related_work}
\paragraph{Face Manipulation Methods.}

In the last two decades interest in virtual face manipulation has rapidly increased.
Breglera~\emph{et al.}~\cite{Bregler1997} presented an image-based approach called Video Rewrite to automatically create a new video of a person with generated mouth movements.
With Video Face Replacement \cite{Dale2011}, Dale~\emph{et al.} presented one of the first automatic face swap methods.
Using single-camera videos, they reconstruct a 3D model of both faces and exploit the corresponding 3D geometry to warp the source face to the target face.
Garrido~\emph{et al.}~\cite{GVRTPT14} presented a similar system that replaces the face of an actor while preserving the original expressions.
VDub~\cite{GVSSVPT15} uses high-quality 3D face capturing techniques to photo-realistically alter the face of an actor to match the mouth movements of a dubber.
Thies~\emph{et al.}~\cite{Thies15} demonstrated the first real-time expression transfer for facial reenactment.
Based on a consumer level RGB-D camera, they reconstruct and track a 3D model of the source and the target actor.
The tracked deformations of the source face are applied to the target face model.
As a final step, they blend the altered face on top of the original target video.
Face2Face, proposed by Thies~\emph{et al.}~\cite{Thies16}, is an advanced real-time facial reenactment system, capable of altering facial movements in commodity video streams, e.g., videos from the internet.
They combine 3D model reconstruction and image-based rendering techniques to generate their output.
The same principle can be also applied in Virtual Reality in combination with eye-tracking and reenactment \cite{thies2018facevr}.

Recently, Suwajanakorn~\emph{et al.}~\cite{suwajanakorn2017synthesizing} learned the mapping between audio and lip motions, while
their compositing approach builds on similar techniques to Face2Face \cite{Thies16}.
Averbuch-Elor~\emph{et al.}~\cite{elor2017bringingPortraits} present a reenactment method, Bringing Portraits to Life, which employs 2D warps to deform the image to match the expressions of a source actor.
They also compare to the Face2Face technique and achieve similar quality.
Other editing by use 3D proxies for 3D object manipulation in a single photograph using stock 3D models \cite{kholgade20143d}, physics-based edits in videos \cite{bazin2016physically,haouchine2017calipso}.

%
Recently, several face image synthesis approaches using deep learning techniques have been proposed.
Lu~\emph{et al.}~\cite{LuLCHS17} provides an overview.
Generative adversarial networks (GANs) are used to apply Face Aging \cite{AntipovBD17}, to generate new viewpoints \cite{HuangZLH17}, or to alter face attributes like skin color \cite{LuTT17}.
Deep Feature Interpolation \cite{UpchurchGBPSW16} shows impressive results on altering face attributes like age, mustache, smiling etc.
Similar results of attribute interpolations are achieved by Fader Networks \cite{LampleZUBDR17}.
Most of these deep learning based image synthesis techniques suffer from low image resolutions.
Karras~\emph{et al.}~\cite{Karras2017} improve the image quality using progressively growing of GANs.
Their results include high-quality synthesis of faces.
%

\paragraph{Multimedia Forensics.}

Multimedia forensics aims to ensure authenticity, origin, and provenance of an image or video
without the help of an embedded security scheme.
Focusing on integrity early methods are driven by handcrafted features
that capture expected statistical or physics-based artifacts that occur during
image formation.
Surveys on these methods can be found in~\cite{Farid16,Sencar13}.
Recently, several CNN-based solutions have been proposed in image forensics~\cite{Bayar16,Cozzolino17,Bondi17:TDL,Bappy2017}.
For videos, the main body of work focuses on detecting manipulations that can
be created with relatively low effort, such as dropped or duplicated
frames~\cite{Wang07:EDF,Gironi14:VFT,Long17:C3D}, varying interpolation
types~\cite{Ding17:IMC}, copy-move manipulations~\cite{Bestagini13:LTD,Cozzolino2018}, or chroma-key
compositions~\cite{Mullan17:RBF}. 
The proposed face benchmark fills this gap in the research landscape by
providing a huge video dataset of advanced synthesized faces.

For forensics specifically on faces, some methods have been proposed to distinguish computer generated faces from natural ones~\cite{Nguyen12:ICG,Conotter14:PBD,Rahmouni2017}, and to detect face retouching~\cite{Bharati16:DFR}.
In biometry, Raghavendra~\emph{et al.}~\cite{Raghavendra17:DeepCNN} recently proposed to detect morphed faces with two pre-trained deep CNNs, VGG19 and AlexNet.
Finally, Zhou~\emph{et al.}~\cite{Zhou17} proposed detection of two different face swapping manipulations using a two-stream network:
one stream detects low-level inconsistencies between image patches while the other stream explicitly detects tampered faces.

However, robustness issues are addressed only in very few works, even though it is of paramount importance for practical applications.
For example, operations like compression and resizing are known for laundering manipulation traces from the data.
Unfortunately, compression and resizing are routinely carried out when images and videos are uploaded to social networks, which is one of the most typical application fields for forensic analysis.
An even greater challenge to a forensic detector are targeted attacks that consist of suitable post-processing steps to hide the traces of manipulation.
All these attacks go under the collective name of counter-forensics~\cite{Bohme12:CFA}.
Forensic analysis and counter-forensics are in continuous competition.
Model-based methods appear to be extremely fragile on laundered data since they focus on specific image features which oftentimes disappear with post-processing.
Data-driven methods can be expected to be more robust, especially if they rely
on data which have a processing history coherent with the asset of
interest~\cite{Boroumand2018}.
A key benefit of the proposed dataset is that its size lifts video forensics
research to a level that allows to create better detectors,
but also better counter-forensics methods on a significant amount of data. 
At the same time, the dataset serves as a unified benchmark.

\paragraph{Datasets.}
Classical forensics datasets have been created with significant manual effort
under very controlled conditions, to isolate specific properties of the data
like camera artifacts.  Most notably, the ``Dresden image database'' consists
of 14,000 images from 73
cameras, and is used primarily for camera fingerprinting \cite{Gloe10:DID}.
The recent VISION dataset also aims at camera fingerprinting, with 34,427 images and 1914 videos that were uploaded and downloaded from social
media~\cite{Shullani17}.

While several datasets were proposed that include image manipulations, only a few of them address also the important case of video.
For image copy-move manipulations a large dataset is
MICC\_F2000 
consisting of a collection of 700 forged images from
various sources~\cite{Amerini11}.
Datasets containing very different and realistic image manipulations are
the First IEEE Image Forensics Challenge Dataset\footnote{http://ifc.recod.ic.unicamp.br/fc.website/index.py?sec=0}, 
which comprises a total of 1176 forged images,
the Wild Web Dataset 
\cite{Zampoglu15} with 90 real cases of manipulations coming from the web
and the Realistic Tampering dataset 
\cite{Korus2016TIFS} including 220 forged images.
    
More recently, Al-Sanjary~\emph{et al.} presented 33 videos on YouTube that
contain different manipulations~\cite{AlSanjary16}. 
The National Institute of Standards and Technology (NIST) presented with the Nimble
Challenge 2017 a large benchmark dataset~\cite{Nimble17}. 
However, it contains a total of 2520 manipulated images, but only 23 manipulated videos with ground truth.
A database of 2010 FaceSwap- and SwapMe-generated images has recently been
proposed by Zhou~\emph{et al.}~\cite{Zhou17}. 
While this dataset is most similar to our proposed benchmark, it is orders of magnitude smaller, and only consists of still images instead of videos.

\section{The FaceForensics Dataset}
\label{sec:dataset}
We introduce the {\em FaceForensics} dataset which is created from \VIDEOS {} videos (i.e., unique identities).
In the following, we describe the data collection and processing used to generate our two datasets.
The first dataset (see \cref{sec:reenactment_db}) contains manipulated videos where the source and target video differs, while the second dataset (see \cref{sec:selfreenactment_db}) consists of videos where Face2Face is used to reproduce the input video (i.e., source and target video are the same).
This second dataset gives us access to ground truth pairs of synthetic and real images.

\paragraph{Data Collection}
The data was collected from YouTube. We chose videos with a resolution larger than 480p that were tagged with "face", "newscaster" or "newsprogram" on the youtube8m dataset \cite{abu2016youtube} as well as other videos that were found on YouTube with these tags.
We use the Viola-Jones \cite{viola2001rapid} face detector to extract video sequences that contain a face for more than 300 consecutive frames.
In addition to that, we perform a manual screening of the resulting clips to ensure a high quality of video selection and to avoid videos with face occlusions.

\paragraph{Data Processing}
To process the video data, we use a variant of the state-of-the-art Face2Face approach \cite{Thies16}, that is able to fully-automatically create reenactment manipulations.
The technique re-renders the face in a target video under possibly different expressions.
We process each video in a preprocessing pass; here, we use the first frames in order to obtain a temporary face identity (i.e., 3D model), and track the expressions over the remaining frames.
In order to improve the identity fitting and the static texture, we select the frames with the left- and rightmost angle of the face in an automated way; in the original implementation of Face2Face this step has to be done manually.
Using these poses, we jointly fit the identity and estimate a static texture.
Based on this identity reconstruction, we track the whole video to compute the per frame expression, rigid pose, and lighting parameters.
The generated tracking and reconstruction results allow us to generate any source-target video combinations for the reenactment.
We generate the reenactment video by transferring the source expression parameters (i.e., 76 Blendshape coefficients) to the target video.
A detailed explanation of the reenactment process can be found in the original paper \cite{Thies16}.
As the result, we store the original source, the target image, and the manipulated output image for each frame.
In addition, we generate a per-pixel binary mask of the modified pixels, which serves as ground truth for segmentation tasks.

\subsection{Source-to-Target Reenactment Dataset}
\label{sec:reenactment_db}
\begin{figure}[t]
\centering
	\includegraphics[width=0.7\linewidth]{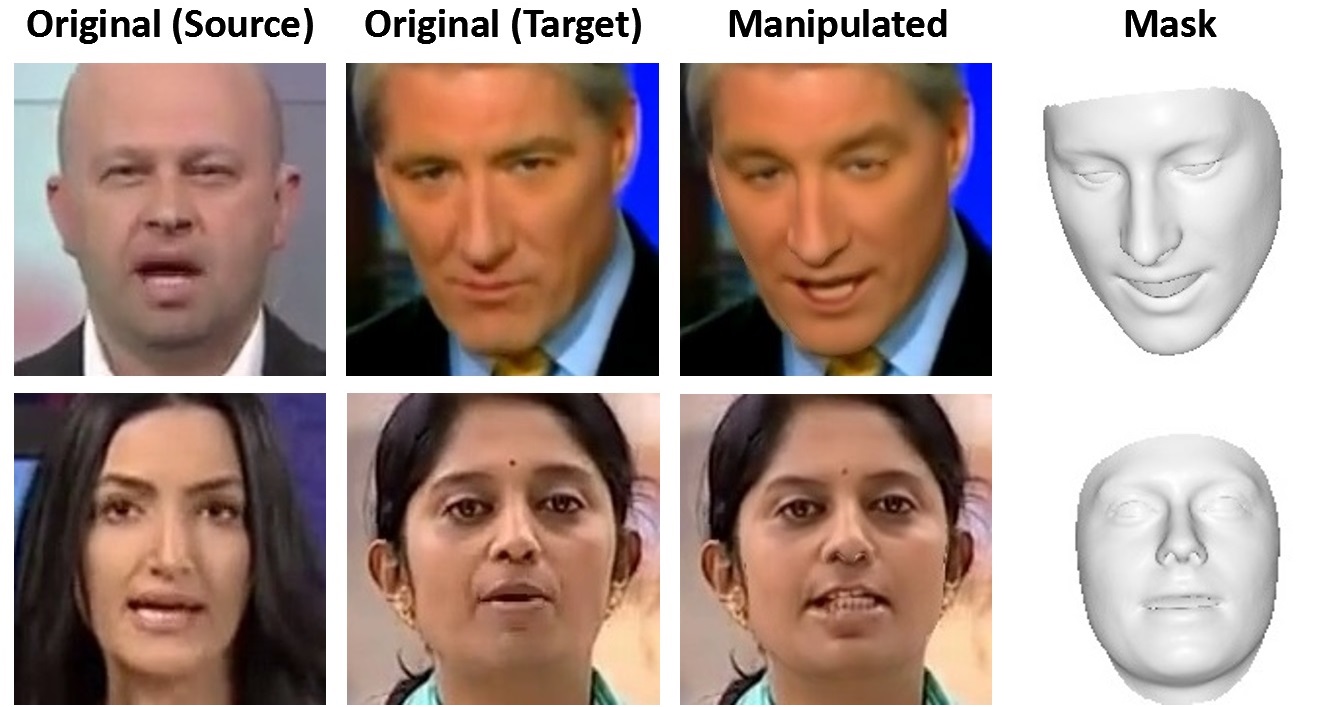}
	\caption{
		Source-to-Target Dataset. From left to right: original input image of the source actor, input image of the target actor, reenactment result and face mask that is used during synthesis of the output image.}
	\label{fig:dataset_selfreenactment_source}
\end{figure}

For the \textit{source-to-target} dataset, we use the original Face2Face reenactment approach between two randomly chosen videos (see \cref{fig:dataset_selfreenactment_source}).
The technique uses a mouth retrieval approach that selects the mouth interiors from a mouth database based on the target expressions.
This person specific mouth database is built upon the tracked videos in the preprocessing step (i.e., contains images of the target video).
The mouth database is one of the most limiting factors of the Face2Face approach, since the videos may not cover a variety of mouth expressions, leading to distortions of the mouth in the resulting reenactment output.
The dataset is split into \NUMTRAIN{} videos for training (\FRAMESTRAIN{} images), \NUMVAL{} videos for validation (\FRAMESVAL{} images), and \NUMTEST{} videos for testing (\FRAMESTEST{} images).
We use the \textit{source-to-target} reenactment dataset for all testing, as well as for training all classification and segmentation approaches; see \cref{sec:detection} and \cref{sec:segmentation}.

\subsection{Self-Reenactment Dataset}
\label{sec:selfreenactment_db}
\begin{figure}[tbh!]
\centering
	\includegraphics[width=0.7\linewidth]{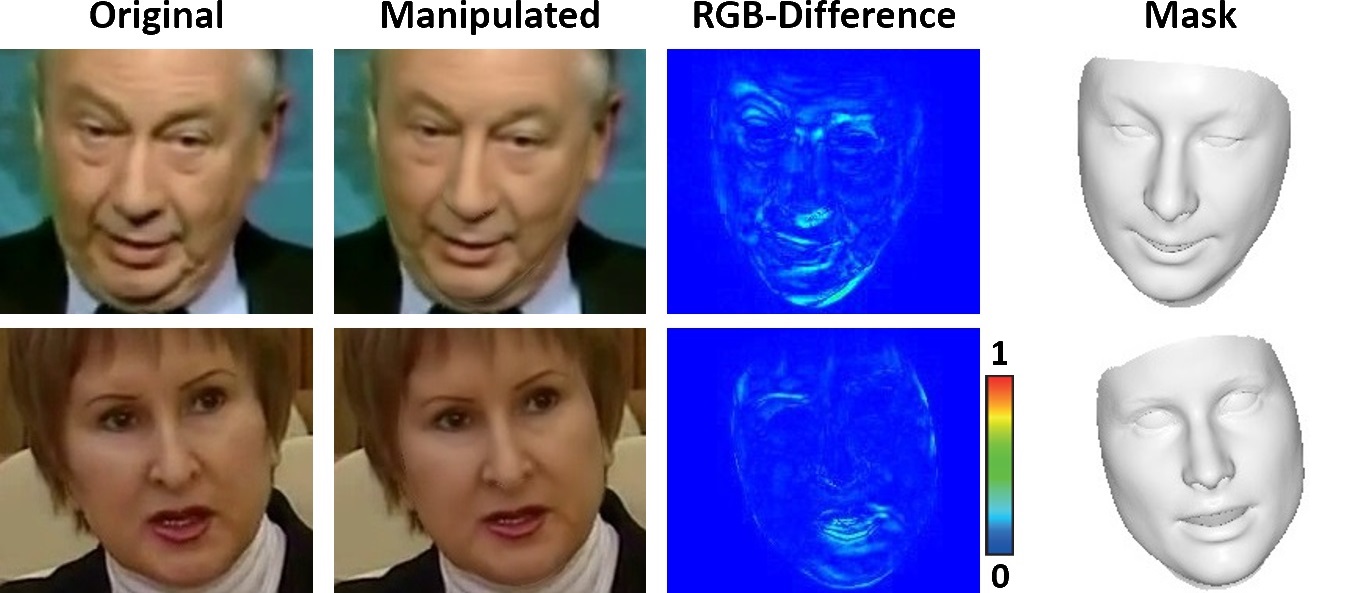}
	\caption{Fake or Real? Examples of the FaceForensics Self-Reenactment Dataset. From left to right: original input image, self-reenacted output image, color difference plot and face mask that is used during synthesis of the output image.
		}
	\label{fig:dataset_selfreenactment_self}
\end{figure}

The second dataset is built upon \textit{self-reenactment} generated by Face2Face (see \cref{fig:dataset_selfreenactment_self}).
Instead of different source and target video combinations, the self reenactment scenario uses the same video as source and target video.
Applying this reenactment technique to a video, we obtain video pairs consisting of ground truth data and manipulated (re-rendered) facial imagery.
These ground truth pairs are ideally suited for training generative approaches for FaceForensics, which we explore in \cref{sec:generative}. 
We split the \textit{self-reenactment} dataset into the same \NUMTRAIN{} videos for training (\FRAMESTRAINSELF{} images), \NUMVAL{} videos for validation (\FRAMESVALSELF{} images), and \NUMTEST{} videos for testing (\FRAMESTESTSELF{} images).

\section{Forgery Classification Task}
\label{sec:detection}
The forgery classification task has the goal to identify forged images.
It is cast as a binary classification problem on a per frame basis of the manipulated videos.
Since there are no specific approaches in the current literature to detect Face2Face manipulations,
we decided to consider learning-based methods used in the forensic community for generic manipulation detection \cite{Bayar16,Cozzolino17},
computer-generated vs natural image detection \cite{Rahmouni2017}
and face tampering detection \cite{Raghavendra17:DeepCNN,Zhou17}.
In addition, we also included a state-of-the-art deep network \cite{Chollet17}.
Each of these methods is trained on the same source-to-target reenactment dataset comprising 10 frames from each of the \NUMTRAIN{} forged and \NUMTRAIN{} pristine videos.
Likewise, the validation and test set both consist of 10 frames extracted from each of \NUMVAL{} (pristine) and \NUMVAL{} (fake) videos. 
For each frame, we crop all images to be centered around the face, where we make use of the face mask provided by Face2Face.
The faces have been resized to the input size of the network when requested \cite{Zhou17,Chollet17},
otherwise, a clip of 128x128 pixels centered on the face has been extracted as input \cite{Bayar16,Cozzolino17,Rahmouni2017}. 

\begin{figure}[t!]
    \centering
	\includegraphics[width=0.8\linewidth]{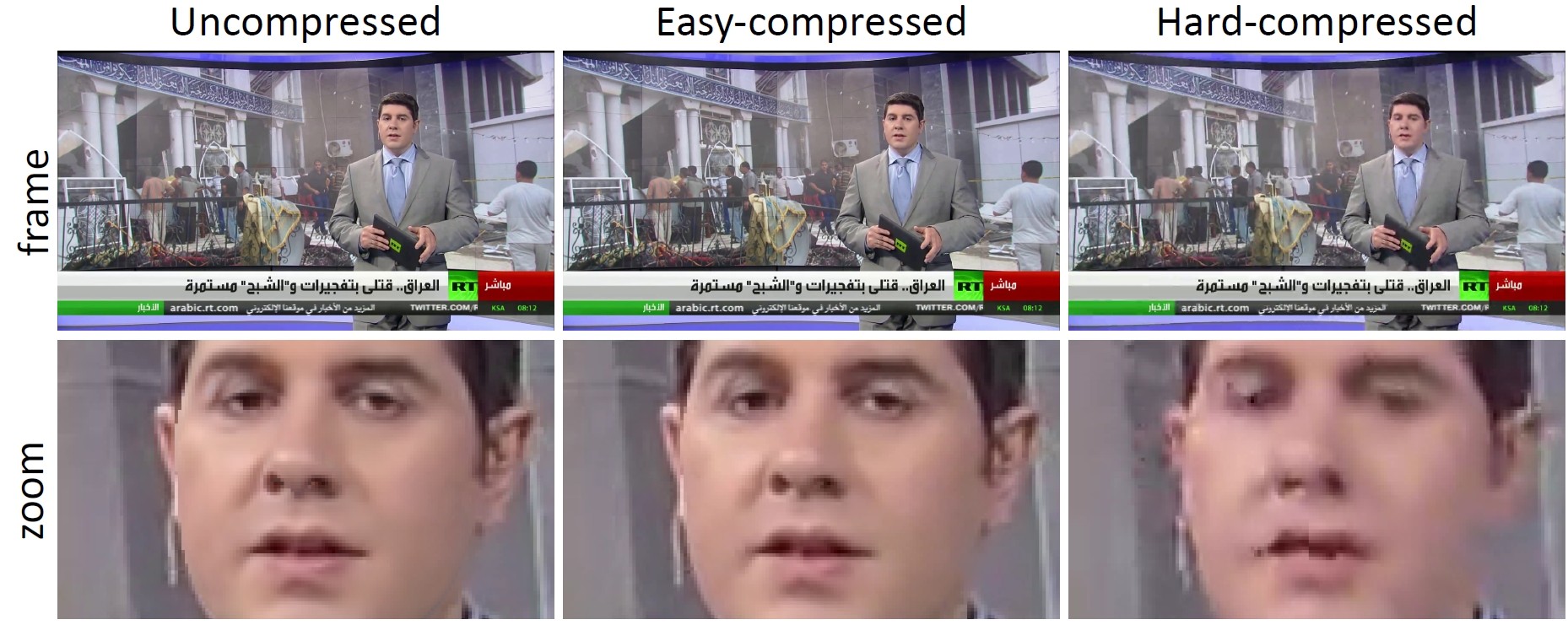}
	\caption{Uncompressed frame (left), easy-compressed (middle), and hard-compressed one (right).}
	\label{fig:dataset_compression}
	\vspace{-0.5cm}
\end{figure}

For all baselines, we evaluate classification accuracy on uncompressed data,
on H.264 compressed data with quantization parameter equal to 23 (light compression) and 40 (strong compression), to cover the quality parameters of a range of different distribution channels, including popular social networks.  
A sample frame extracted from these three settings is shown in Fig.~\ref{fig:dataset_compression}.
In the following, we briefly describe all the approaches used for comparison.

{\em Steganalysis Features + SVM}:
it is a handcrafted solution based on the extraction of co-occurrences on 4 pixels patterns along the horizontal and vertical direction on the high-pass images, proposed originally in steganalysis~\cite{Fridrich12:RMS},
using only one single model (for a total feature length of 162) 
which was the winning approach in the first IEEE Image forensic Challenge~\cite{Cozzolino14:Phase1}. 
These features are then used to train a linear SVM classifier.

{\em Cozzolino et al.} 2017 \cite{Cozzolino17}: use a CNN-based network implementing
the handcrafted features described above. 
The network is then fine-tuned on our dataset.
{\em Bayar and Stamm} 2016 \cite{Bayar16}: propose a CNN-based network with 8 layers: a constrained convolutional layer, 2 convolutional layers, 2 max-pooling layers and 3 fully-connected layers. The constrained convolutional layer is specifically designed to suppress the high-level content of the image.

{\em Rahmouni et al.} 2017 \cite{Rahmouni2017}: adopt different CNN architectures with a global pooling layer that computes four statistics (mean, variance, maximum and minimum).
We consider the network that had the best performance (Stats-2L).
{\em Raghavendra et al.} 2017 \cite{Raghavendra17:DeepCNN}: use two pre-trained CNNs VGG19 and AlexNet.
The networks are fine-tuned on our dataset, then the feature vectors extracted from the first fully connected layer of the two networks are concatenated and used as input for the Probabilistic Collaborative Representation Classifier.
{\em Zhou et al.} 2017 \cite{Zhou17}: 
consider a two-stream network, a pre-trained deep CNN, fine-tuned on our dataset,
(GoogLeNet Inception V3 model) 
and a patch triplet stream trained on 5514D steganalysis features~\cite{Goljan15:CFA}. 
The final score is then obtained by combining the output scores of the two streams. 
 
In addition to these approaches, we also evaluate a transfer learning model of the state-of-the-art XceptionNet CNN architecture~\cite{Chollet17}.
It is based on depthwise separable convolution layers with residual connections.
XceptionNet is pre-trained on ImageNet
and fine-tuned on our source-to-target reenactment dataset. 
During fine-tuning, we freeze the first 36 layers which corresponds to the first 4 blocks of the network.
Only the last layer is replaced by a dense layer with two outputs, initialized randomly and trained anew for 10 epochs. 
After that, we train the resulting network until the validation does not change in 5 consecutive epochs. 
For optimization, we use the following hyperparameters for our reported scores: ADAM \cite{kingma2014adam} with a learning rate of $0.001$, $\beta_1=0.9$ and $\beta_2=0.999$ as well as a batch-size of $64$.
	
In Tab.~\ref{tab:classification}, we show a comparison of these methods applied to
uncompressed and compressed videos. In the absence of compression, all methods, including \cite{Fridrich12:RMS} based on handcrafted features, achieve a relatively high performance.
For compressed videos, performance drops, particularly for handcrafted features and for shallow CNN architectures \cite{Bayar16,Cozzolino17}. 
Deep neural networks are better at handling these situations, with XceptionNet slightly outperforming the method by Zhou~\emph{et al.}~\cite{Zhou17}. 
On the other hand, even humans have a hard time detecting manipulations under strong compression as shown in \cref{refinementvisual}.

\newcommand{\ru}    {\rule{0mm}{3.8mm}}
\begin{table}[t!]
	\begin{center}
		\begin{tabular}{|l|c|c|c|} \hline
			\ru	Methods                                       & ~~~no-c~~~  & ~~~easy-c~~~ & ~~~hard-c~~~ \\ \hline \hline
			\ru	\cite{Fridrich12:RMS}~Steganalysis Features + SVM~            & 99.40 & 75.87 & 58.16 \\ \hline
			\ru	\cite{Cozzolino17}~Cozzolino~\emph{et al.}~	            & 99.60 & 79.80 & 55.77 \\ \hline 
			\ru	\cite{Bayar16}~Bayar and Stamm	                        & 99.53 & 86.10 & 73.63 \\ \hline
			\ru \cite{Rahmouni2017}~Rahmouni~\emph{et al.}~             & 98.60 & 88.50 & 61.50 \\ \hline
			\ru \cite{Raghavendra17:DeepCNN}~Raghavendra~\emph{et al.}~ & 97.70 & 93.50 & 82.13 \\ \hline
			\ru	\cite{Zhou17}~Zhou~\emph{et al.}                        & 99.93 & 96.00 & 86.83 \\ \hline
			\ru	\cite{Chollet17}~XceptionNet                            & 99.93 & 98.13 & 87.81 \\ \hline
		\end{tabular}
	\end{center}
    \vspace{1mm}
	\caption{Classification accuracy (face-level detection; i.e., is a face manipulated or not) of reference methods with no compression ({\em no-c}), light compression ({\em easy-c}), and strong compression ({\em hard-c}) using our FaceForensics benchmark dataset.}
	\label{tab:classification}
\end{table}

\section{Forgery Segmentation Task}
\label{sec:segmentation}

Pixel-level segmentation of manipulated images (also referred to as forgery localization in the forensics community) is a very challenging task.
The most successful approaches proposed in the image forensics literature rely on camera-based artifacts (e.g. sensor noise, demosaicking).
However, their application on the frames extracted from our dataset did not provide satisfactory results, not even for uncompressed data.
Hence, we discard them and focus only on deep learning methods, which can take full advantage of our dataset for training.
In particular, those proposed in \cite{Cozzolino17} and \cite{Rahmouni2017} already perform localization and need no further adaptation.

Additionally, considering its very good performance in classification, we adapt also XceptionNet~\cite{Chollet17} to the localization task, as described in the following.

At test time, the network runs in sliding-window modality on patches of $128\times 128$ pixels, with stride 16.
For each patch, $W_i$, it outputs the estimated manipulation probability, $\hat{p}_i = {\rm CNN}(W_i)$, which is assigned to the central $16\times 16$ region.

Preliminary to training, a ground truth is computed by labeling as manipulated all pixels that have been modified with respect to the pristine frame.
Spurious pixels are removed by morphological filtering, and a spatial filtering is performed to smooth boundaries.
Eventually, the ground truth pixels range from 0 (pristine background) to 1 (manipulated face), with intermediate values on the boundaries,
and such values are regarded as manipulation probabilities $p_i$.
These will be used to compute the loss function as the cross-entropy between ground-truth and estimated probabilities
$\sum_i -(p_i)\log(\hat{p}_i) -(1-p_i)\log(1-\hat{p}_i)$,
where the sum goes over all patches of a mini-batch, and $p_i$ is the ground truth probability of the central pixel.

The patch-level training set is formed by taking 10 frames from each training set video,
and 3 patches from each frame, one from the face, one from the background, and one over the face-background boundary.
Training is performed using ADAM, with mini-batches of 96 patches,
formed by taking the 3 (pristine) plus 3 (fake) patches associated with 16 forged frames and with the corresponding 16 pristine frames.
For each epoch, the frames are shuffled, preserving the correspondence between pristine and forged patches.
We use a learning rate of $0.0001$, $\beta_1=0.9$, $\beta_2=0.999$, batch-size $96$, and again train the resulting network until the validation does not change in 5 consecutive epochs.

\begin{figure*}
	\centering 
	\includegraphics[width=.30\linewidth]{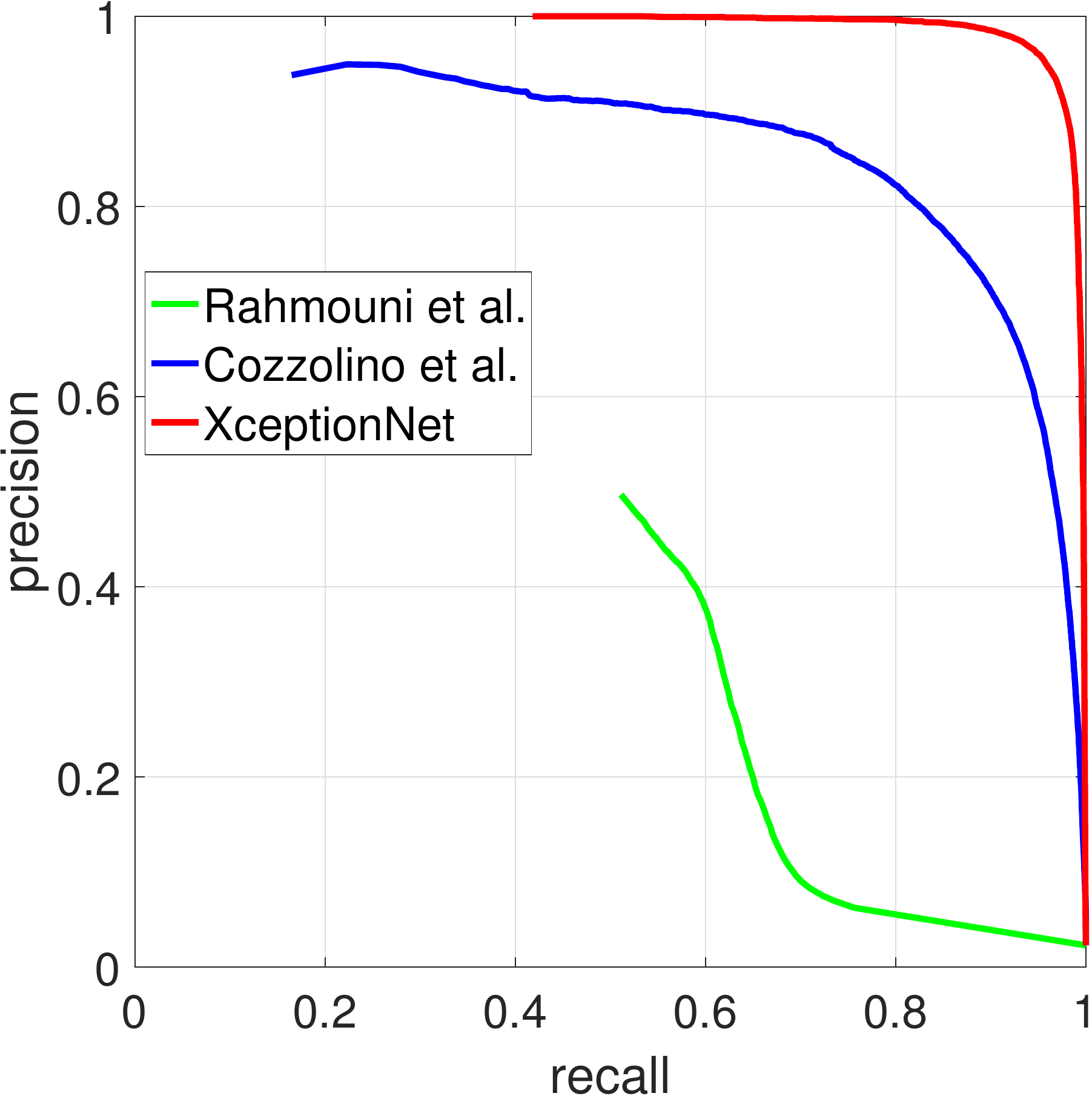}
	\hspace{0.25cm}
	\includegraphics[width=.30\linewidth]{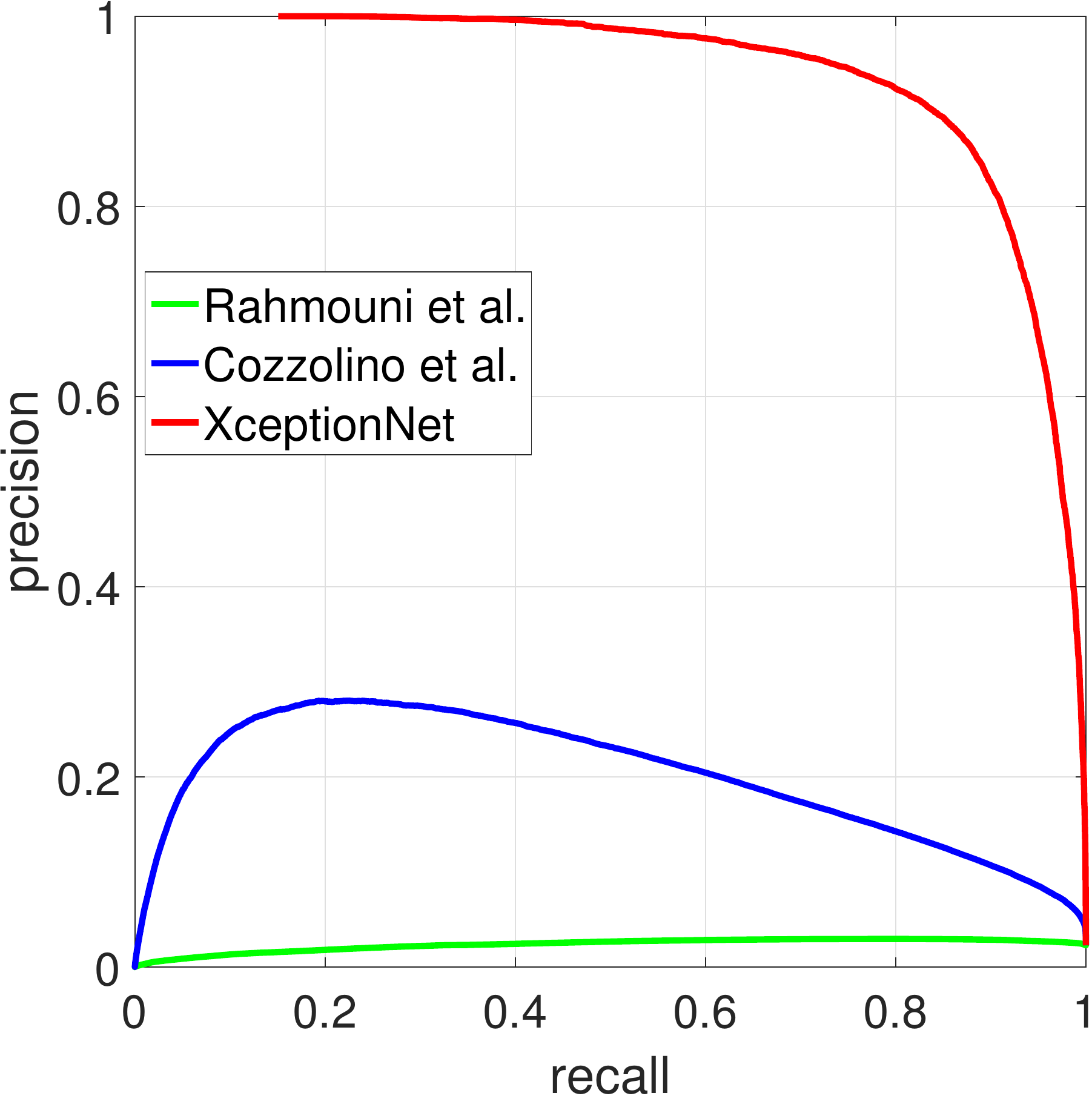}
	\hspace{0.25cm}
	\includegraphics[width=.30\linewidth]{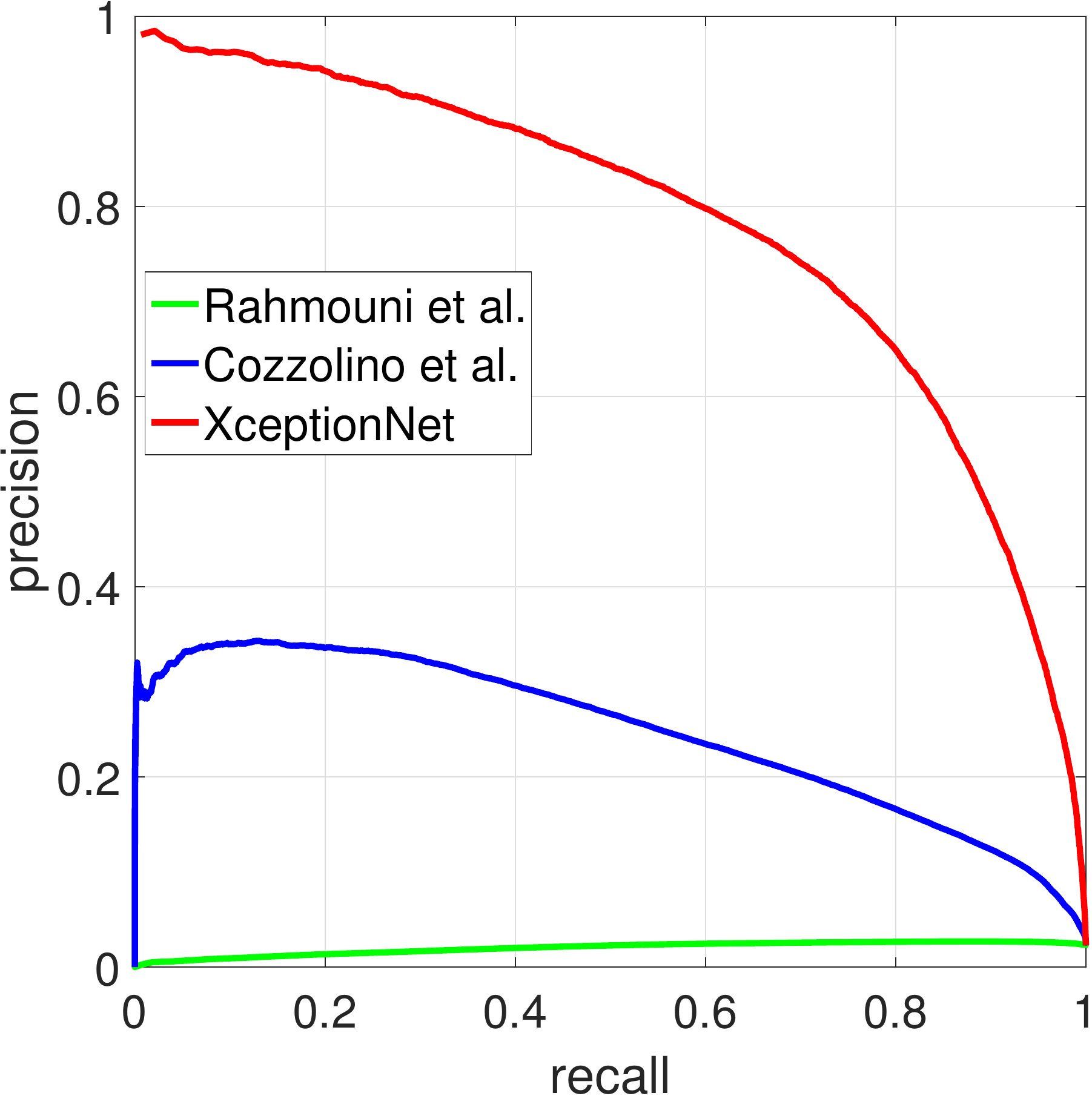}
	\caption{
		Precision vs recall on uncompressed videos (left), easy-compressed (middle), and hard-compressed (right).
		The test set comprises both forged and pristine images.
	}
	\label{fig:ROCs1}
\end{figure*}

\begin{figure*}[t!]
	\centering
    \includegraphics[width=\linewidth]{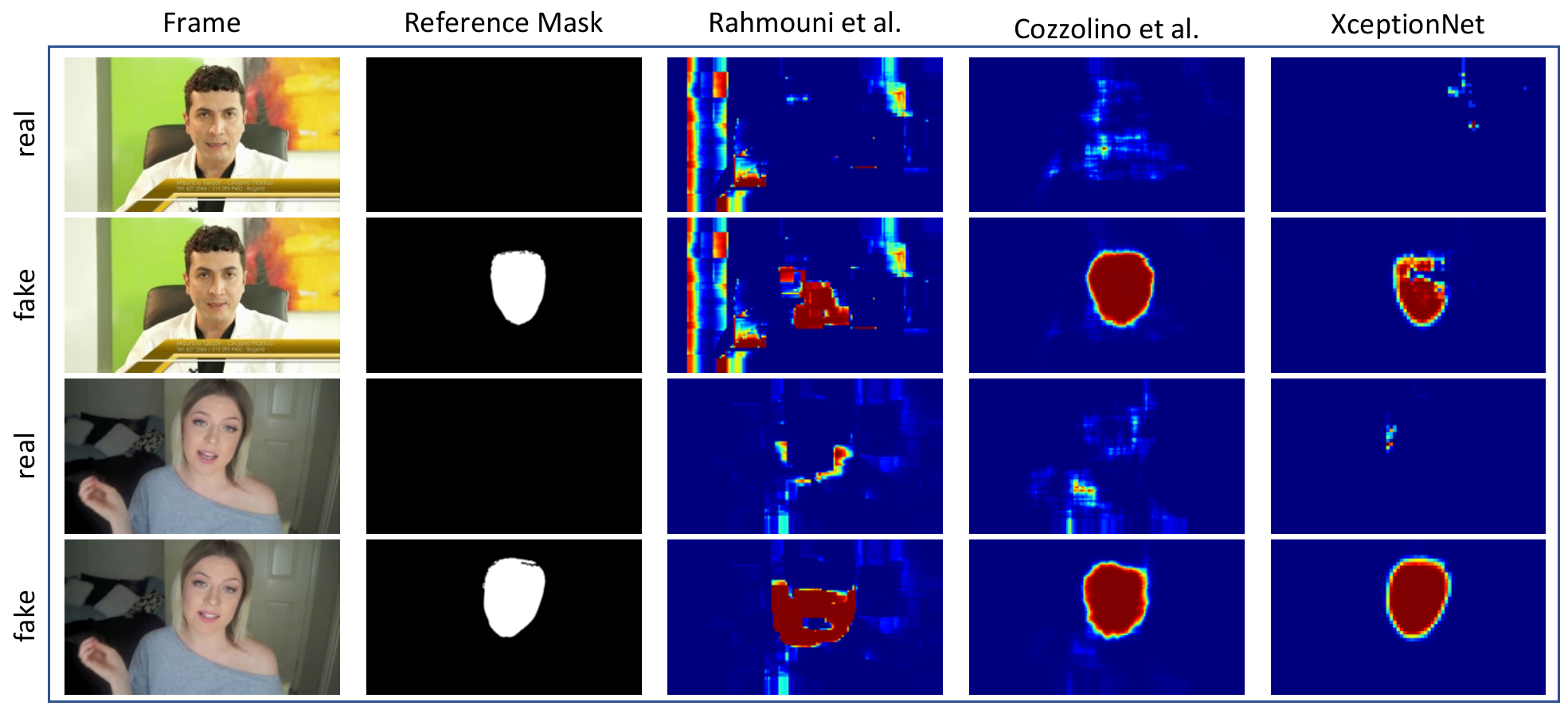}
	\caption{
        Forgery segmentation results.
		For each frame, we show the heatmaps for the original video (first row) and the manipulated one (second row).
		From left to right: input frame, ground truth mask (only for the fake input), results of Rahmouni~\emph{et al.}~\cite{Rahmouni2017},  Cozzolino~\emph{et al.}~\cite{Cozzolino17}, and the XceptionNet-based method.
		Both Cozzolino~\emph{et al.} and XceptionNet reliably localize the manipulations on uncompressed data.
	}
	\label{fig:segmentation}
\end{figure*}

\begin{figure*}
\centering
\includegraphics[width=\linewidth,trim=0 122 0 0, clip]{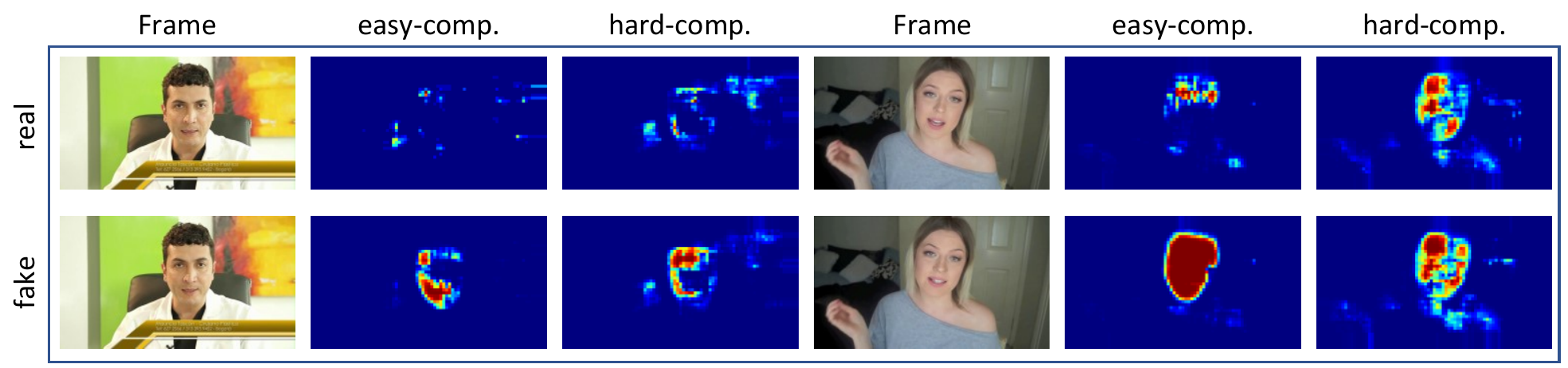}
\caption{Forgery segmentation vs. compression. 
    With increasing compression rate, the segmentation results get worse. 
    The XceptionNet-based method is still able to segment the manipulation in most cases, even under hard compression.}
\label{fig:segmentation_compression}
\end{figure*}

In \cref{fig:ROCs1}, we show a quantitative evaluation of these methods.
For all of them, performance degrades with increasing compression rate, as more and more false positives and false negatives occur.
Eventually, at the highest compression rate, only the method based on XceptionNet keeps providing good results.
In \cref{fig:segmentation} and \cref{fig:segmentation_compression}, we also show visual results over both uncompressed and compressed data.
In this last case, we only show results provided by XceptionNet, since the other two methods output useless heatmaps.

\section{Refinement Task}
\label{sec:generative}
\Cref{sec:detection} shows that Face2Face manipulations can be detected quite easily in an uncompressed setting with a sufficiently large amount of data.
This gives rise to the question whether such an amount of data can also be used in the opposite direction, i.e., to improve the quality of the manipulations.
To this end, we leverage the self-reenactment dataset which contains \FRAMESSELF{} manipulated frames with target ground truth pairs for supervised training. 

\begin{figure*}[t]
	\centering
	\includegraphics[width=1.0\linewidth]{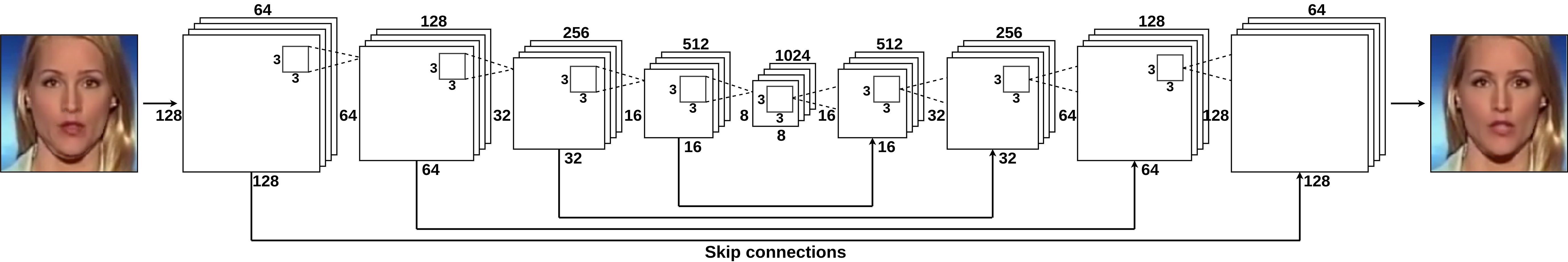}
	\caption{The autoencoder (AE) architecture with skip connections used for refining the forged images. The AE is first pre-trained on the significantly larger, but unlabeled, VGGFace2 dataset in an unsupervised fashion (w/o enabling the skip connections). We then fine tune on our self-reenactment training set using supervision with Face2Face and target ground truth training pairs.}
	\label{fig:autoencoder}
\end{figure*}

As a baseline, we devise an autoencoder CNN architecture with skip connections that takes as input a $128\times 128$ pixels image and predicts an image of the same resolution (see Fig.~\ref{fig:autoencoder} for the detailed architecture).
In order to obtain meaningful and strong features for images of human faces, we first pre-train the autoencoder network on a self-reconstruction task using the VGGFace2 dataset \cite{cao2017vggface2} in an unsupervised fashion.
This dataset contains $3.31$ million images of $9131$ subjects, which is about an order of magnitude more than our dataset but does not provide annotations.
In this pre-training process, where we use all of these images for training, we disable the skip connections, thus forcing the networks to solely rely on the bottleneck layer.
We then fine tune the pre-trained autoencoder network on our FaceForensic self-reenactment dataset using the \FRAMESTRAINSELF{} training images.
Here, we input the manipulated faces and constrain it with the known target ground truth using an $\ell_1$ loss in the supervised training process; note that we aim to minimize the difference image, which is a widely-used technique.
In addition, we enable the skip connections which allow us to obtain sharper results in the autoencoder output.
At test time, we feed in data from the FaceForensic source-to-target test dataset in order to improve the quality of forgeries.
We optimize the network with ADAM using a batch size of 32, a learning rate of $0.001$, $\beta_1=0.9$, $\beta_2=0.999$, and continue training until convergence on the self-reenactment validation set.

The main advantage of an autoencoder architecture over a network without a bottleneck layer is the ability to leverage the larger (unlabeled) VGGFace2 dataset for unsupervised pre-training.

\begin{figure}
	\centering
	\includegraphics[width=0.7\linewidth]{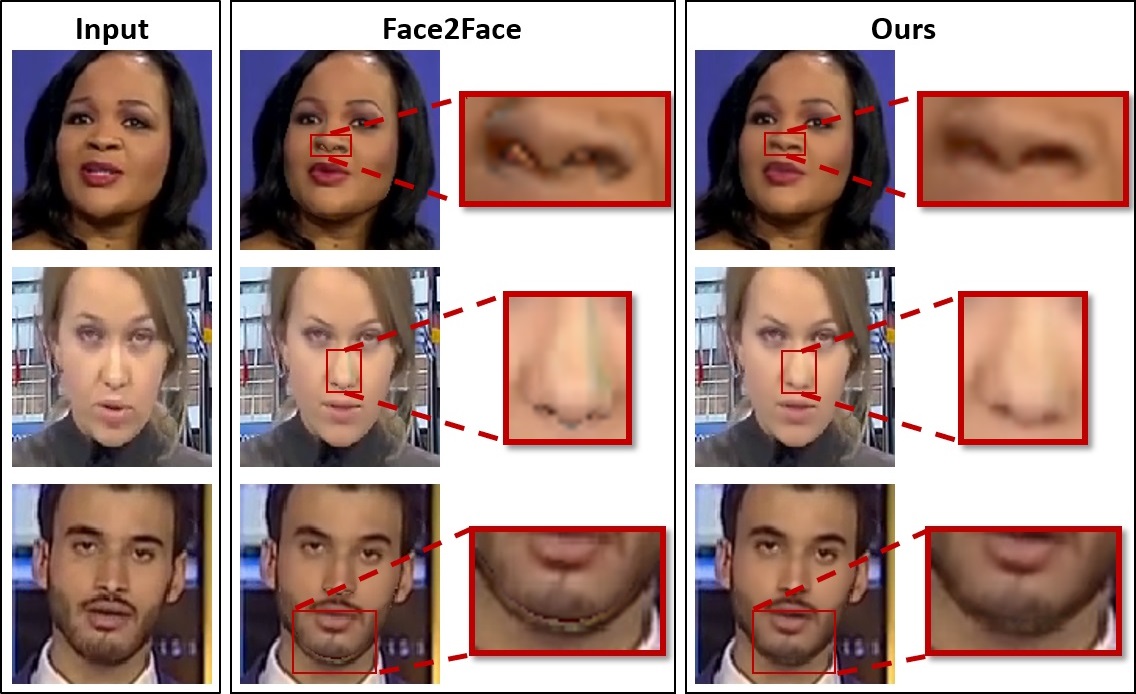}
	\caption{
		Refinement of our autoencoder approach:
		we can see in the close-ups that our refinement significantly improves the visual result of Face2Face~\cite{Thies16}.
		Especially, regions around the nose, the chin, and the cheek, where most of the artifacts of the Face2Face method occur, are corrected.
		Errors occurring in the transition between the mask region, hosting the reconstructed and modified face, and the unmodified background are removed by our method.
		The autoencoder also improves regions where a wrong illumination estimate in the Face2Face algorithm leads to artifacts (e.g., see second row).
	}
	\label{fig:refinement}
\end{figure}

\subsection{Perceptual Evaluation of the Refinement}\label{refinementvisual}
Fig.~\ref{fig:refinement} shows a qualitative comparison of the Face2Face reenactment approach and our refinement results using our source-to-target test dataset.
It shows that most artifacts of Face2Face are visible in the area of the chin, the cheek, and the nose.
These are border regions of the face mask that is used to re-synthesize the face and the original image.
The autoencoder significantly improves these regions and nicely blends between 
foreground manipulated face, delimited by the mask region, with the 
target video stream in the background.
Illumination errors in the Face2Face output are also corrected by our method.

In order to compare the visual quality of forged images obtained with Face2Face and our refinement network, we conduct a user study with 14 participants, whose results are shown in Tab.~\ref{tab:user_study}.
The 14 participants are Master and Ph.D. students in computer science who are not involved in this project.
For the study, we randomly choose 50 images from each of the no-compression and easy-compression, and 20 from the hard-compression categories.
All images are taken from the source-to-target test set at a $128^2$ resolution, and we select images at a ratio of 50\% pristine and 50\% forged; i.e., we have 25 fake and 25 pristine images for the no-c and easy-c categories, and 10 fake and 10 pristine images on the hard-c one.
For each participant, we randomly shuffle the images within each of the compression categories.
Before showing an image from a category, we let the participants know which category is being presented; i.e., explain details regarding compression.
We show each image for three seconds to a participant, then the image goes blank, and either real or fake has to be chosen.
This process is repeated for each image and for each compression category.
We conduct this experiment for the raw Face2Face output \cite{Thies16} and for our refined results obtained with the autoencoder network.

Quantitative results show that that humans are worse at identifying manipulations than the XceptionNet-based approach.
For highly-compressed images, this becomes particularly obvious, as human accuracy is about 50\%, which is essentially random guessing.
For the easier compression setups, participants are able to identify better than random chance; however, accuracy is still relatively low.
We can also clearly observe that our autoencoder refinement makes visual differences even harder to spot, thus increasing the quality of forgeries for human observers.

	\begin{table}[t!]
	\begin{center}	
		\begin{tabular}{|l|c|c|c|} \hline
			\ru	User Study			  		& ~~~no-c~~~  & ~~~easy-c~~~ & ~~~hard-c~~~ 	\\ \hline \hline
			\ru	{\bf Face2Face: w/o AE}         & 68.71 & 62.00  & 50.00 \\	\hline
			\ru	{\bf Face2Face: Refined w/ AE~~~} 	& 60.57 & 51.29  & 48.93 \\	\hline
		\end{tabular}
    \vspace{3mm}    
		\caption{User study results on images generated with the raw Face2Face output (top)	and with the proposed refinement approach (bottom). Candidates are shown an image for three seconds and have to classify it into real or fake. We see that it is noticeably harder for humans to identify forgeries after refinement.}
		\label{tab:user_study}
	\end{center}
\end{table}

\subsection{Quantitative Evaluation}

However, we can also evaluate our refiner with the classification methods described in \cref{sec:detection}.  As we aim to improve the quality of our fakes, the created data should be more difficult to detect than without refinement under the same circumstances, namely identical classifier architecture and amount of training data. Therefore, we use the same evaluation protocol as in \cref{sec:detection}, i.e., we refine 10 images for every video in our source-to-target training, validation and test set. In addition to that, we resize face images to $128\times 128$ pixels for a fair comparison between the refined and raw images and retrain XceptionNet on the resulting dataset as in \cref{sec:detection}. 

\begin{table}
	\begin{center}
		\begin{tabular}{|l|c|c|c|} \hline
			\ru	Datasets on 128x128			  		& ~~~no-c~~~  & ~~~easy-c~~~ & ~~~hard-c~~~ 	\\ \hline \hline
			\ru	{\bf Face2Face: w/o AE}         & 99.42 & 96.17  & 84.56 \\	\hline
			\ru	{\bf Face2Face: Refined w/ AE~~~} 	& 99.23 & 96.07  &  80.97 \\	\hline
		\end{tabular}
	\end{center}
	\caption{Classification accuracy of a XceptionNet on our source-to-target test dataset using images of $128\times 128$ pixels. In the top row, we show results on the fake data directly generated by Face2Face; in the second row, we use our autoencoder refiner that is trained on the self-reenactment training set. The autoencoder succeeds to slightly lower detection performances under strong compression.}
\label{tab:refinement_results}
\end{table}

In \cref{tab:refinement_results} we observe that the autoencoder slightly lowers the detection accuracy on compressed data, but it does not affect the overall performance by a large margin.
Therefore, even if the visual quality of fakes seems to be high, there are still many shortcomings that make these methods easy to detect for forgery detection algorithms as the classifier is still able to detect refined fakes with high accuracy, which suggests that visual results itself seem to be a poor metric.
One possibility to circumvent this problem and produce high quality refinements would be generative adversarial networks \cite{goodfellow2014generative}, which have already been successfully applied to unsupervised refinement \cite{shrivastava2017learning} and were shown to be able to produce high-resolution results \cite{Karras2017}.

\section{Conclusions}
	
In this work, we introduce a novel dataset of manipulated videos that exceeds all existing publicly available forensic datasets by orders of magnitude. 
We provide a benchmark for general image forensic tasks on this dataset such as identification and segmentation of forged images. 
We show that handcrafted approaches are highly challenged by realistic amounts of compression, whereas we set a strong baseline of results for detecting a facial manipulation with modern deep learning architectures. 

We also introduce a second application of the dataset, by visually improving the quality of the forgery with an autoencoder that is trained in a supervised fashion on our self-reenactment dataset. 
However, our refiner mainly improves visual quality, but it only slightly encumbers forgery detection for deep learning method trained exactly on the forged output data. 
This motivates us to further investigate refinement methods in future work, as we believe that this interplay between tampering and detection is not only an extremely exciting avenue for follow-up work but also of utmost importance in order to build robust and generalizable classifiers.

\section{Acknowledgement}
We gratefully acknowledge the support of this research by the AI Foundation, a TUM-IAS Rudolf M\"o{\ss}bauer Fellowship, and Google Faculty Award.
In addition, this material is based on research sponsored by the Air Force Research Laboratory and the Defense Advanced Research Projects Agency under agreement number FA8750-16-2-0204. 
The U.S. Government is authorized to reproduce and distribute reprints for Governmental purposes notwithstanding any copyright notation thereon. The views and conclusions contained herein are those of the authors and should not be interpreted as necessarily representing the official policies or endorsements, either expressed or implied, of the Air Force Research Laboratory and the Defense Advanced Research Projects Agency or the U.S. Government.

\section*{\LARGE Appendix}
\begin{appendix}
\section{User Study Interface}

\begin{figure*}
	\centering
	\includegraphics[width=0.5\linewidth]{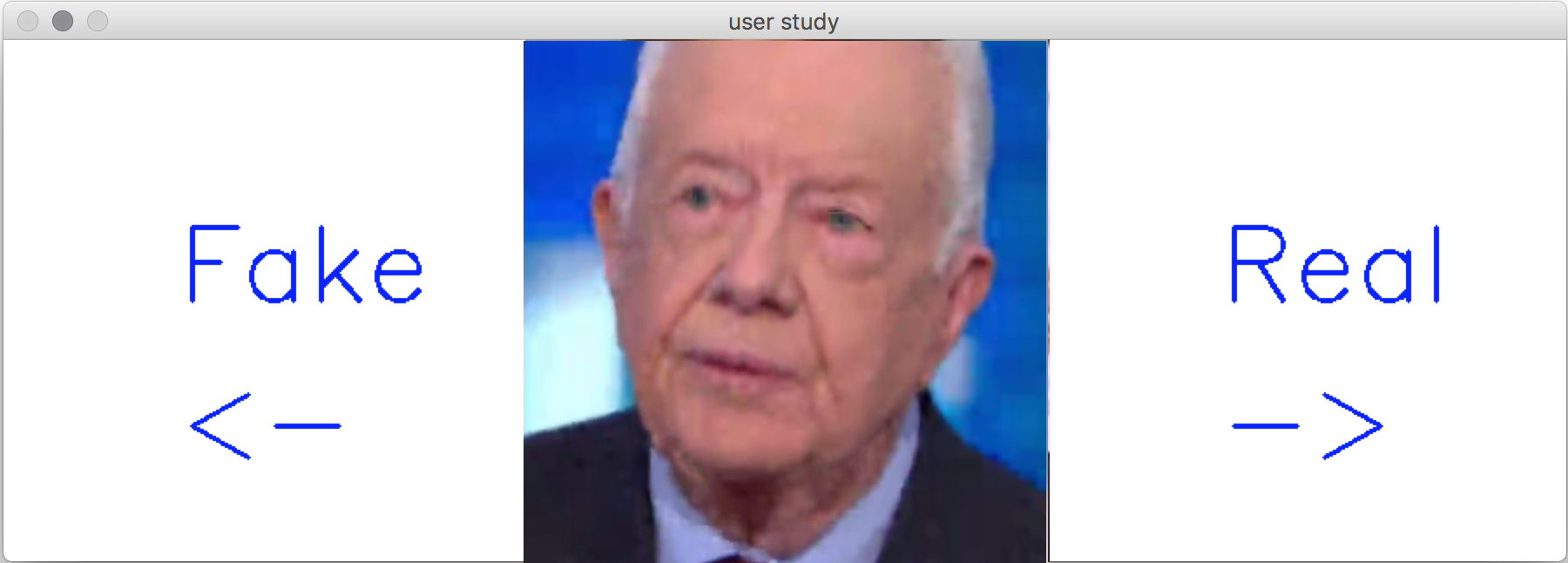}
	\includegraphics[width=0.5\linewidth]{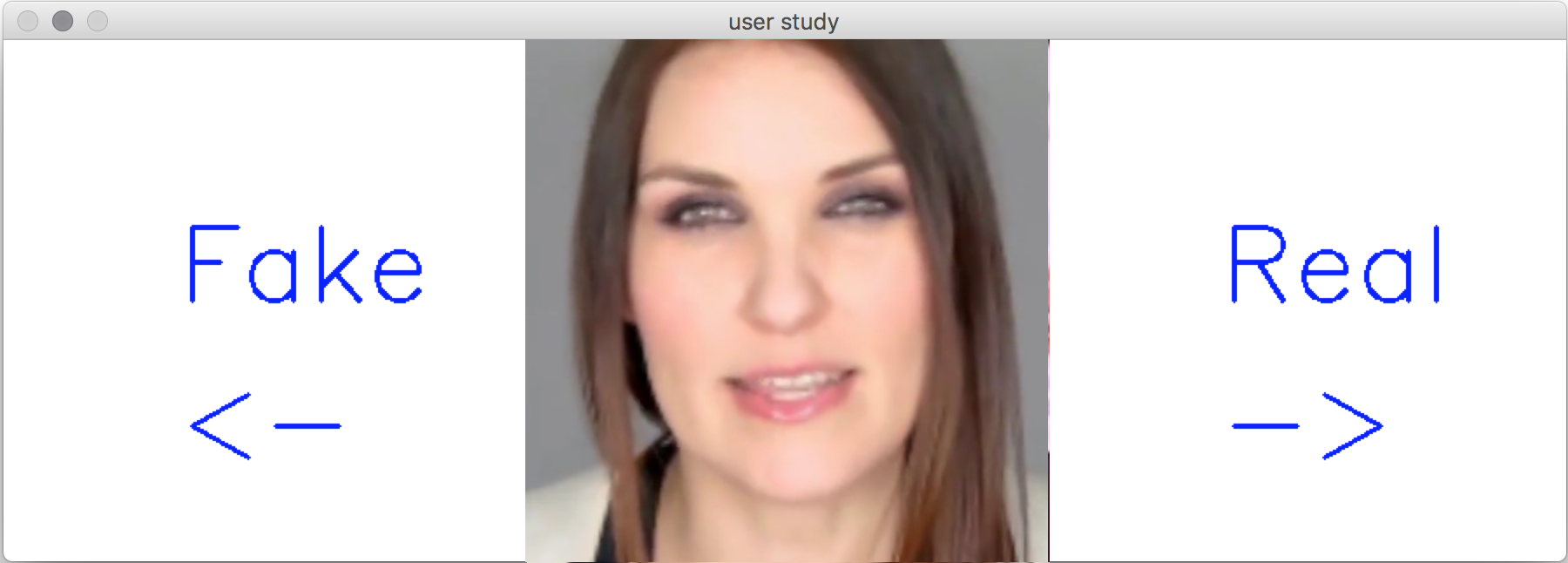}
	\includegraphics[width=0.5\linewidth]{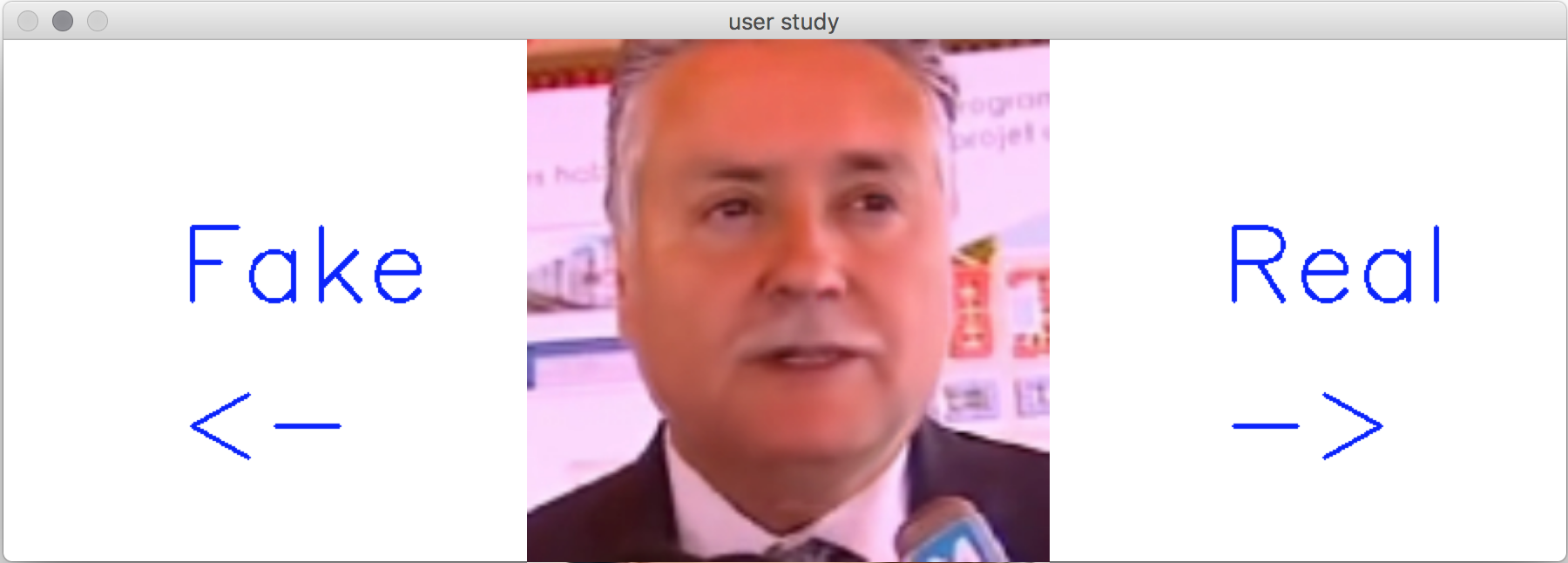}
	\caption{Three exampes of our user interface; we show each image for three seconds, after the participant needs to select either fake or real. By the way, the top row is real (hard compression), the second row is fake (easy compression), the bottom row is fake with refinement (uncompressed).}
	\label{fig:userstudy}
\end{figure*}

Fig.~\ref{fig:userstudy} visualizes the interface for conducting our user study. We show participants a random images, and ask them to select fake or real. The images are randomly chosen from the no-, easy-, and hard-compression sets; however, we ensure an equal number of fake an real images for each participant.

\section{Forgery Segmentation Examples}

In this section, we show additional qualitative results on forgery segmentation 
for compressed and uncompressed videos for three CNN-based architectures:
Rahmouni et al. \cite{Rahmouni2017}, Cozzolino et al. \cite{Cozzolino17}
and XceptionNet \cite{Chollet17}.

\begin{figure*}[htb!]
	\centering
	\includegraphics[width=\linewidth,trim=0 4 0 0, page=2]{img/loc.pdf} 
	\includegraphics[width=\linewidth,trim=0 0 0 20, clip, page=3]{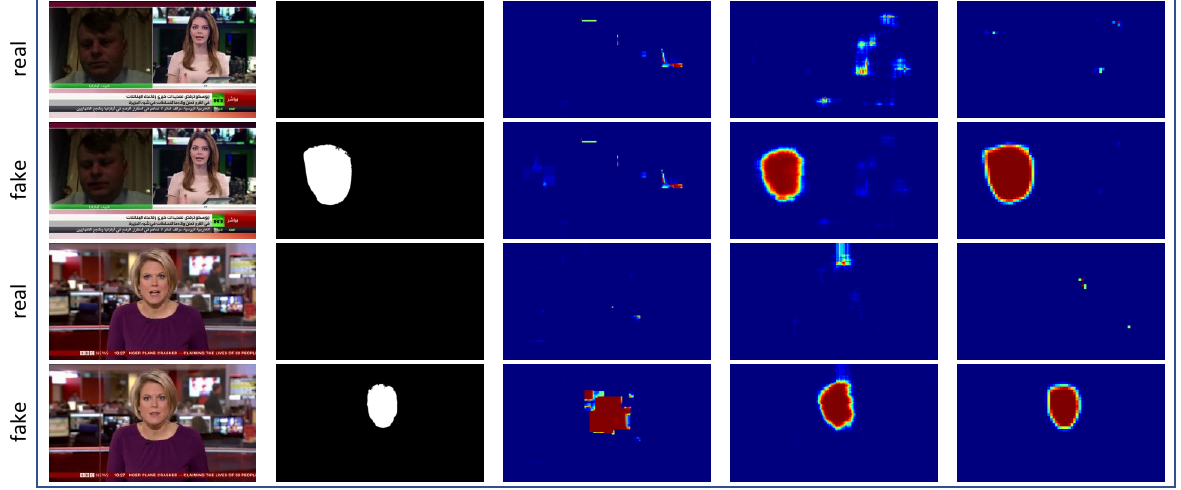}
	\caption{
		Additional forgery segmentation examples.
		For each frame, we show the heatmaps for the original video (first row) and the manipulated one (second row).
		From left to right: input frame, ground truth mask (only for the fake input), results of Rahmouni~\emph{et al.}~\cite{Rahmouni2017},  Cozzolino~\emph{et al.}~\cite{Cozzolino17}, and the XceptionNet-based method.
	}
	\label{fig:segmentation_supplemental}
\end{figure*}

Fig.~\ref{fig:segmentation_supplemental} shows results on umcompressed video where we can see that the XceptionNet model provides the best results compared to \cite{Rahmouni2017} and \cite{Cozzolino17}.
It is able to correctly locate the manipulated area on the fake videos,
while on real videos, we can hardly notice any false positives.

\begin{figure*}[htb!]
	\centering
	\includegraphics[width=\linewidth,trim=0 4 0 0, page=1]{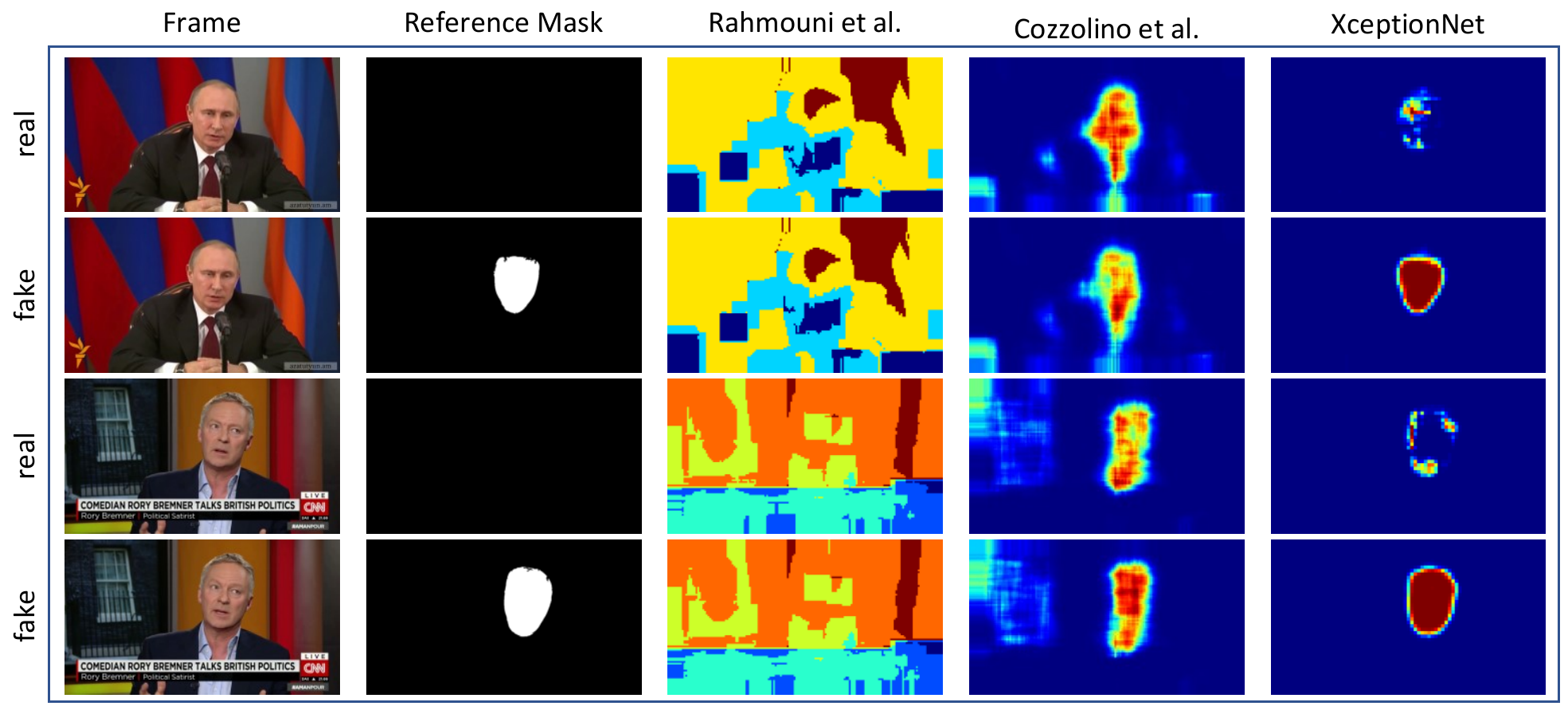} 
	\includegraphics[width=\linewidth,trim=0 0 0 20, clip, page=2]{img/loc_easy.pdf}
	\caption{
		Forgery segmentation examples on easy-compressed videos.
		For each frame, we show the heatmaps for the original video (first row) and the manipulated one (second row).
		From left to right: input frame, ground truth mask (only for the fake input), results of Rahmouni~\emph{et al.}~\cite{Rahmouni2017},  Cozzolino~\emph{et al.}~\cite{Cozzolino17}, and the XceptionNet-based method.
	}
	\label{fig:segmentation_supplemental_easy}
\end{figure*}

\begin{figure*}[htb!]
	\centering
	\includegraphics[width=\linewidth,trim=0 4 0 0, page=1]{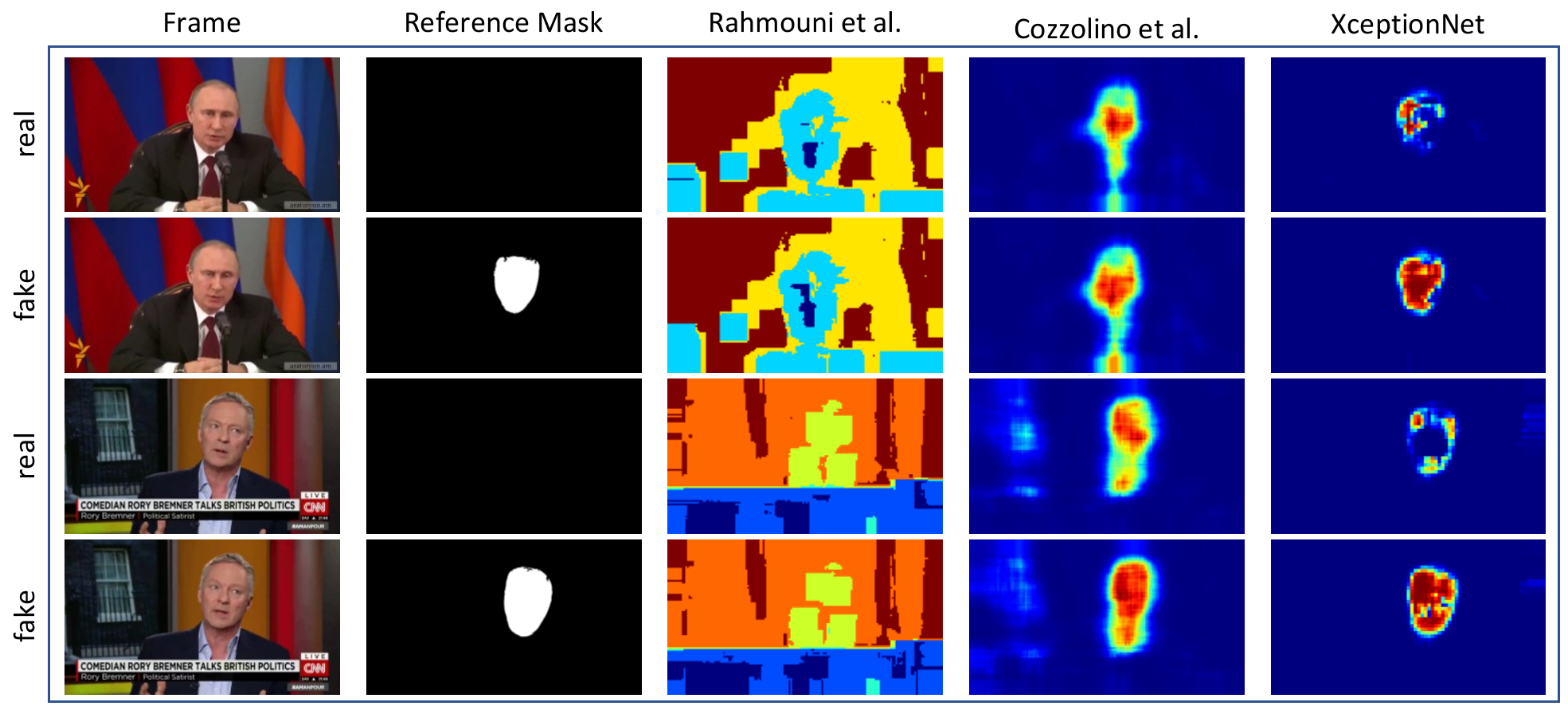} 
	\includegraphics[width=\linewidth,trim=0 0 0 20, clip, page=2]{img/loc_hard.pdf}
	\caption{
		Forgery segmentation examples on hard-compressed videos.
		For each frame, we show the heatmaps for the original video (first row) and the manipulated one (second row).
		From left to right: input frame, ground truth mask (only for the fake input), results of Rahmouni~\emph{et al.}~\cite{Rahmouni2017},  Cozzolino~\emph{et al.}~\cite{Cozzolino17}, and the XceptionNet-based method.
	}
	\label{fig:segmentation_supplemental_hard}
\end{figure*}

On compressed videos, segmentation becomes more difficult.
The methods proposed by Rahmouni et al. \cite{Rahmouni2017} and Cozzolino et al. \cite{Cozzolino17} produce almost random heat\-maps,
while the XceptionNet model provides results at acceptable quality. 
In Fig.~\ref{fig:segmentation_supplemental_easy}
we can see that with easy-compressed videos XceptionNet still works pretty well allowing reliable segmentation of altered pixels.
If we again increase the compression rate, the task becomes even more challenging, and the resulting segmentation is rather poor, but much better than the compared methods (see Fig.~\ref{fig:segmentation_supplemental_hard}).

\end{appendix}

\bibliographystyle{splncs}
\bibliography{egbib}

\end{document}